\documentclass[12pt]{article}
\usepackage{amsmath}
\usepackage{amssymb}
\usepackage{times}
\usepackage{graphicx}
\usepackage{color}
\usepackage{multirow}
\usepackage[authoryear]{natbib}
\usepackage{rotating}
\usepackage{bbm}
\usepackage{latexsym}

\newcommand{\captionfonts}{\normalsize}

\makeatletter
\long\def\@makecaption#1#2{%
  \vskip\abovecaptionskip
  \sbox\@tempboxa{{\captionfonts #1: #2}}%
  \ifdim \wd\@tempboxa >\hsize
    {\captionfonts #1: #2\par}
  \else
    \hbox to\hsize{\hfil\box\@tempboxa\hfil}%
  \fi
  \vskip\belowcaptionskip}
\makeatother

\begin{document}
\hspace{13.9cm}1

\ \vspace{20mm}\\

{\LARGE Binary Sparse Codes: Where are we on Energy Efficient Representations?}

\ \\
{\bf \large Jose Miguel Ramos$^{\displaystyle 1}$, Luis SaCouto$^{\displaystyle 2}$, Andreas Wichert$^{\displaystyle 1}$}\\
{$^{\displaystyle 1}$Instituto Superior T\'{e}cnico, Universidade de Lisboa, Portugal}\\
{$^{\displaystyle 2}$Quinnipiac University, USA}\\


\thispagestyle{empty}
\markboth{}{Binary Sparse Codes}

\ \vspace{-0mm}\\

\begin{center} {\bf Abstract} \end{center}
Biological neural systems achieve impressive computational capabilities under tight metabolic constraints, relying on sparse cortical representations that minimise total neural activity. This combination of efficiency and representational power has motivated growing interest in sparse coding as a computational strategy for artificial systems. Artificial models designed to produce binary sparse codes span three broad paradigms, theoretical methods grounded in efficient coding, deep learning methods trained via backpropagation, and biologically plausible methods that learn through local Hebbian rules, each yielding codes whose quality depends not only on achieving low activation density but also on preserving the informational structure of the input. The latter property, however, lacks an agreed-upon quantitative definition, leaving the field without a principled basis for comparing methods. We propose a set of metrics that formalise information preservation as similarity preservation, measuring how faithfully a binary sparse code reproduces the relational geometry of the input space, and apply this framework to compare four representative methods under matched conditions. The results reveal a consistent trade-off in which models that achieve high information preservation do so at the cost of sparsity, while models approaching the uniformity of an optimal random sparse code tend to sacrifice it. Competitive Sparse Codes, a biologically plausible approach, is the only architecture to occupy a region of near-maximal sparsity while retaining meaningful information preservation. These findings frame the simultaneous satisfaction of both factors as an open challenge for the field and provide a reusable evaluation framework for measuring progress toward it.

% ----------------------------------------------------------------
\section{Introduction}
% ----------------------------------------------------------------

The past decade has witnessed remarkable progress in artificial intelligence, with deep neural networks achieving human or superhuman performance across vision, language, and control tasks \citep{lecun2015deep}, while requiring power budgets that dwarf the roughly twenty watts consumed by the human brain \citep{raichle2002appraising}. Biological evolution treats energy consumption as a hard constraint on which computational strategies can survive, and under this constraint biological intelligence relies on sparse cortical representations \citep{rolls1995sparseness} and local synaptic learning rules \citep{hebb1949organization}, properties that together reduce metabolic cost by minimising total neural activity and avoid the coordination overhead of backpropagation by allowing each synaptic change to depend only on locally available activity. The dominant deep learning paradigm relies instead on dense representations and global learning rules. This architectural and energy consumption difference suggests that sparse coding and local learning may be key ingredients in the search for more energy-efficient alternatives.

Beyond energy efficiency, sparse codes, and binary sparse codes in particular, carry an independent theoretical justification in associative memory models \citep{willshaw1969nonholographic}, where storage capacity is maximised precisely when the fraction of active units is small \citep{palm1980associative}, giving sparsity a theoretical grounding that holds independently of any biological argument. These converging motivations have produced a growing body of work seeking computational models capable of generating high-quality binary sparse codes, with approaches spanning theoretical methods grounded in sparse approximation and signal processing \citep{olshausen1996emergence}, deep learning methods trained end-to-end via backpropagation \citep{jang2017categorical}, and biologically plausible methods that learn through local, activity-dependent synaptic rules \citep{sacouto2023competitive,ravichandran2020learning,ravichandran2021brainlike,knoblauch2020iterative}, each differing sharply in learning mechanism yet sharing the same underlying goal of producing codes whose quality depends on more than sparsity alone. A high-quality sparse code must satisfy two competing factors, the sparsity and uniformity of firing across the population and the preservation of the input's information, of which sparsity admits a precise and established definition for binary codes because the fraction of active units is unambiguous, while information preservation admits no comparable standard, since no single agreed-upon way exists to quantify how much of an input's information a representation retains.

Sparsification can be understood as a form of compression even when the code space is higher-dimensional than the input, given that only a small number of its dimensions may be active at once, and compression pushed far enough is bound to sacrifice information as fewer effective degrees of freedom must account for the same variety of inputs, leading to the hypothesis that increasing sparsity comes at a growing cost to information preservation. Achieving high sparsity alone is straightforward when information preservation is not required, just as preserving most of the input's information is straightforward when sparsity is not required, but satisfying both factors simultaneously proves considerably harder. This paper proposes a set of metrics designed to quantify how faithfully a binary sparse code preserves the information of its input and applies this framework to compare the existing families of sparse coding models under matched conditions.

Among the methods evaluated, the biologically plausible approach is the only one able to achieve a firing distribution comparable to that of a random sparse code without completely sacrificing information preservation in the process, though the degree of preservation it attains remains lower than that achieved by methods which do not reach that level of sparsity, confirming that the problem of satisfying both factors simultaneously and robustly has not been perfectly solved by any existing approach. This work is therefore best understood not as identifying a definitive solution but as formulating an open challenge for the field, namely finding sparse coding architectures that satisfy both quality factors simultaneously. The evaluation framework introduced here provides the methodological foundation on which this challenge can be addressed and progress reliably measured. The central contribution is an investigation into the current state of sparse binary codes, establishing where existing methods stand, what they are capable of achieving, and how their quality can be measured and compared.

% ----------------------------------------------------------------
\section{Related Work}
% ----------------------------------------------------------------

\subsection{Sparse Coding}
\label{sec:sparse-coding}

Several independent arguments converge on sparse coding as a promising representational strategy, ranging from direct biological evidence to theoretical arguments from associative memory, efficient coding, and energy efficiency.

Direct biological evidence that the brain operates using sparse codes has been gathered empirically, with sparse neuronal responses measured in both the inferior temporal cortex \citep{rolls1995sparseness} and primary visual cortex \citep{vinje2000sparse}, and the principle extends to simpler organisms \citep{dasgupta2017neural}.

A second, independent argument comes from the theory of associative memory, in which Willshaw, Buneman \& Longuet-Higgins \citep{willshaw1969nonholographic} proposed a simple model where pairs of binary patterns are stored by strengthening synaptic connections between co-active neurons, with a partial cue later reactivating the stored pattern through Hebbian recall, and Palm \citep{palm1980associative} subsequently showed that the storage capacity of such a network, defined as the maximum number of patterns reliably retrievable, is maximised when the number of active units scales as approximately $\log_2(n)$, a regime known as logarithmic sparsity. This result gives sparse coding a theoretical grounding independent of any biological argument, since a network that stores associations optimally must use sparse rather than dense codes. This conclusion also emerges independently in other domains such as vector quantization for nearest-neighbour search \citep{jegou2011product} and discrete representation learning \citep{vanderoord2017neural}.

Another independent argument comes from the efficient coding hypothesis, first proposed by Barlow \citep{barlow1959sensory}, who argued that the retina is structured specifically to remove redundancy from visual input before it is transmitted, so that a limited number of nerve fibres can carry as much genuinely new information as possible, an argument Barlow \citep{barlow1961possible} subsequently generalised beyond the retina to propose that sensory relays throughout the nervous system recode their input specifically to reduce redundancy. Atick \& Redlich \citep{atick1990towards,atick1992retina} extended this redundancy-reduction argument to a noisy channel, deriving the filter that maximises information transmission under a fixed capacity constraint and showing that the predicted filter interpolates between full decorrelation at high signal-to-noise and smoothing at low signal-to-noise, matching how real retinal receptive fields change shape with light level. This decorrelation-based account, however, addresses only redundancy arising from pairwise correlations, whereas natural images also contain higher-order statistical dependencies, most notably the coordinated structure of oriented edges, that a purely pairwise decorrelating code cannot capture, leading Field \citep{field1994goal} to argue that exploiting this structure requires a sparse code in which each unit is silent for most inputs and fires strongly only when its preferred structure is present, a criterion that natural images specifically favour and that set up the coding scheme Olshausen \& Field \citep{olshausen1996emergence} went on to demonstrate directly on natural image patches.

A further, related argument connects sparse coding directly to metabolic cost, since generating a neural spike consumes a disproportionate share of a neuron's energy budget, making a code that keeps most units silent at any moment intrinsically cheaper than one that keeps many units firing simultaneously. Laughlin, de Ruyter van Steveninck \& Anderson \citep{laughlin1998metabolic} estimated this cost directly, finding that transmitting a single bit of information costs on the order of $10^4$ ATP molecules at a chemical synapse and $10^6$--$10^7$ ATP molecules for spike-based coding, several orders of magnitude above the thermodynamic minimum, and Attwell \& Laughlin \citep{attwell2001energy} used this estimate to construct an energy budget for cortical signalling, concluding that the brain's available energy supply can support no more than around 15\% of neurons being simultaneously active. Lennie \citep{lennie2003cost} reached an even stricter estimate, arguing that the high cost of a single spike limits sustained concurrent activity to well under 1\% of cortical neurons, consistent with measured mean firing rates of less than one spike per second per neuron, so that under this budget sparse activity is not merely a theoretically convenient coding strategy but a metabolic necessity, reinforcing from an energetic standpoint the same conclusion that the efficient coding and associative memory arguments arrive at independently.

\subsection{Sparse Coding Methods}

This work compares one representative method from each of three main paradigms for generating sparse codes, a theoretical or mathematical approach, a deep learning approach, and a biologically plausible approach, alongside an untrained baseline, selecting Dictionary Learning (DL) \citep{olshausen1996emergence} for the theoretical approach because it is the direct computational result of the efficient coding argument developed in Section~\ref{sec:sparse-coding}, the Gumbel Autoencoder (GAE) \citep{jang2017categorical} for the deep learning approach as the most suitable method for generating binary sparse codes within a backpropagation framework, Competitive Sparse Codes (CSC) \citep{sacouto2023competitive} for the biologically plausible approach as it learns through local, Hebbian-style learning rules rather than global backpropagation, and Random Projections (RP) \citep{johnson1984extensions} as an additional untrained baseline.

\subsubsection{Dictionary Learning}

Dictionary Learning, introduced by Olshausen \& Field \citep{olshausen1996emergence}, is the foundational method for learning sparse representations from natural images, in which an input signal $\mathbf{x} \in \mathbb{R}^n$ is mapped onto a sparse code $\mathbf{z} \in \mathbb{R}^m$ together with a dictionary $\mathbf{D} \in \mathbb{R}^{n \times m}$ such that $\mathbf{x} \approx \mathbf{D}\mathbf{z}$, obtained by minimising

\begin{equation}
  \label{eq:dl-objective}
  \|\mathbf{x} - \mathbf{D}\mathbf{z}\|^2 + \lambda\|\mathbf{z}\|_1
\end{equation}

through an optimisation that alternates between a sparse coding step, in which $\mathbf{z}$ is found given $\mathbf{D}$, and a dictionary update step, in which $\mathbf{D}$ is updated given $\mathbf{z}$. As motivated in Section~\ref{sec:sparse-coding}, training this objective on natural image patches yields dictionary atoms that converge to spatially localised, oriented, bandpass filters closely resembling V1 simple cell receptive fields, the first computational demonstration that sparse coding emerges directly from the statistics of natural images.

Despite this biological relevance, Dictionary Learning requires global optimisation with no locality constraint and scales poorly with image size, since inference requires solving an iterative optimisation problem for every input rather than a single feed-forward pass, a cost that grows with both input dimensionality and dictionary size \citep{gregor2010learning}, and unlike the other methods evaluated in this work it produces continuous-valued codes rather than binary ones. To enable a fair comparison under matched sparsity conditions, we therefore apply a $k$-winners-take-all ($k$-WTA) operation as a post-processing step, retaining only the $k$ largest activations and setting the rest to zero, which yields binary sparse codes of the desired density, with the drawback that the information preservation scores reflect both what was learned in continuous space and the distortion introduced by binarisation.

\subsubsection{Autoencoders with Gumbel-Softmax Activation}

Autoencoders learn compressed representations by training an encoder-decoder pair to minimise the reconstruction error between an input and its reconstruction, but in their standard form the bottleneck representation is continuous and dense, making autoencoders ill-suited as sparse coding models without explicit modification. Earlier deep learning approaches to this problem include $k$-sparse autoencoders \citep{makhzani2013ksparse}, which enforce sparsity by zeroing all but the top-$k$ hidden activations during both training and inference, and winner-take-all autoencoders \citep{makhzani2015wta}, which use a lifetime sparsity constraint across the batch to prevent individual units going dead. Jang et al.\ \citep{jang2017categorical} proposed addressing this by replacing the bottleneck with a Gumbel-Softmax function, partitioning the code into $N$ groups of $K$ cells and treating each group as an independent categorical variable relaxed via a softmax perturbed by Gumbel noise, with a hard one-hot winner taken per group through the straight-through estimator at encoding time. This construction yields codes that are exactly sparse by design, with exactly $N$ active bits out of $N \times K$ total, one per group, requiring no separate thresholding step since the group structure itself enforces the target sparsity level. However, this structural guarantee does not ensure that the winning cell is distributed equally across all $K$ options within each group over the dataset, leaving open the possibility that a small subset of cells consistently wins while the rest go almost entirely unused, an outcome that would drastically reduce the effective code dimensionality despite the nominal sparsity being maintained. To enforce equal usage across all cells within each group, a categorical KL-divergence penalty is applied per group over each training batch, pushing the average winner distribution toward uniform. Rather than adding the KL term to a scalar loss with a fixed weight, the two gradient contributions are balanced in norm at every step.

The effective weight is computed per batch as $\beta = \lambda \cdot \|\nabla_Z \mathcal{L}_{\text{recon}}\| / \|\nabla_Z \mathcal{L}_{\text{KL}}\|$, so that the combined gradient

\begin{equation}
  \label{eq:gae-gradient}
  \nabla_Z \mathcal{L} = \nabla_Z \mathcal{L}_{\text{recon}} + \beta \cdot \nabla_Z \mathcal{L}_{\text{KL}}
\end{equation}

has reconstruction and sparsity contributions of equal norm when $\lambda = 1$, with $\lambda$ scaling that ratio to control the relative importance of code uniformity against reconstruction fidelity, where
\begin{equation*}
  \mathcal{L}_{\text{KL}} = \sum_{g=1}^{N} \mathrm{KL}(\bar{q}_g \| u),
\end{equation*}
$\bar{q}_g = (\bar{\pi}_{g,1}, \ldots, \bar{\pi}_{g,K})$ is the average softmax distribution across the batch for group $g$, $u = (1/K, \ldots, 1/K)$ is the uniform distribution over $K$ cells, and $\mathrm{KL}(\bar{q}_g \| u) = \sum_{k=1}^{K} \bar{\pi}_{g,k} \log(K\,\bar{\pi}_{g,k})$. The effect of $\lambda$ is studied in the experiments that follow, where configurations spanning a range of values reveal a predictable trade-off: lower values allow the model to prioritise reconstruction fidelity, producing codes with higher information preservation at the cost of less uniform bit usage across the population, while higher values enforce tighter uniformity at the cost of reconstruction quality. Together, these choices allow the encoder to be trained end-to-end via backpropagation despite the categorical, discrete nature of the bottleneck, though unlike biologically plausible models' local Hebbian rule this requires global backpropagation through the full encoder-decoder architecture, making it computationally more expensive than locally-learned alternatives.

\subsubsection{Competitive Sparse Codes}

The Competitive Sparse Codes algorithm, proposed by Sa-Couto \& Wichert \citep{sacouto2023competitive}, is a biologically plausible network designed to encode images into sparse binary codes suitable for associative memory, motivated directly by the capacity analysis in Section~\ref{sec:sparse-coding}, in which achieving the logarithmic sparsity required by Willshaw's network requires a learned encoder capable of mapping high-dimensional natural images onto sparse binary codes while preserving their similarity structure.

The architecture is organised into $N$ groups of neurons, each assigned to a local receptive field, a region of the input image within which the group independently learns $K$ prototype vectors representing the typical patterns observed in that region. During encoding, the neurons within each group compete to produce a single winner through a $k$-WTA mechanism, yielding exactly one active bit per group, so that the final sparse code, formed by concatenating the outputs of all groups, is a binary vector of length $N \times K$ containing exactly $N$ active bits.

The placement of receptive fields is a design choice not fixed by the algorithm and one that may have a substantial effect on what structure each group is able to learn, since groups whose receptive fields cover uninformative or near-empty regions of the input will produce less discriminative prototypes than those covering regions with richer variation. We therefore examine two strategies for assigning receptive field positions in the experiments that follow: one in which positions are drawn uniformly at random, following the configuration of the original paper (CSC-R), and one in which positions are drawn from a two-dimensional Gaussian centred on the image (CSC-G), making groups more likely to cover the informationally richer central region of the digit and less likely to be assigned to peripheral regions that are typically background.

Hebb \citep{hebb1949organization} proposed that the synaptic connection between two neurons strengthens in proportion to the correlation between their activities, commonly summarised as ``cells that fire together wire together,'' and Competitive Sparse Codes implements this principle directly, with weight updates within each group depending only on the correlation between the receptive field's input and the activity of that group's winning neuron, with no information flowing between groups and no global loss function guiding the update. This strict locality makes the algorithm biologically plausible and computationally lightweight, since the total number of trainable parameters scales linearly with the number of groups and training requires only a single pass through the data per epoch.

\subsubsection{Random Projections}

Random Projections offer a learning-free baseline grounded in the geometry of high-dimensional spaces, in which an input $\mathbf{x} \in \mathbb{R}^n$ is encoded by projecting it through a fixed random matrix $\mathbf{R} \in \mathbb{R}^{NK \times n}$, set at initialisation and never updated. The projection is partitioned into $N$ groups of $K$ units, and within each group a winner-take-all operation selects the single largest activation, yielding a one-hot binary vector of length $K$ per group. The final code is formed by concatenating these $N$ group codes,

\begin{equation}
  \label{eq:rp-code}
  \mathbf{z} = \bigl[\,\mathrm{WTA}(\mathbf{R}_1\mathbf{x}),\; \ldots,\; \mathrm{WTA}(\mathbf{R}_N\mathbf{x})\,\bigr]
\end{equation}

where $\mathbf{R}_g \in \mathbb{R}^{K \times n}$ is the $g$-th block of $\mathbf{R}$, producing a binary vector of length $N \times K$ with exactly $N$ active bits, one per group. This group structure matches that of the other methods in the comparison, ensuring that the sparsity regime is identical across all four. The theoretical justification for retaining geometric structure derives from the Johnson--Lindenstrauss property \citep{johnson1984extensions}, according to which random projections approximately preserve pairwise distances with high probability, suggesting that the input space geometry is retained even in the absence of optimisation. Because the projection is blind to the data distribution, however, the resulting codes carry no learned structure, which positions Random Projections as a lower bound in the comparison, a method that is maximally efficient yet makes no use of the input statistics.

% ----------------------------------------------------------------
\section{Evaluation Metrics}
% ----------------------------------------------------------------

\subsection{Sparsity}
\label{sec:sparsity}

The sparse codes considered throughout this paper are binary vectors produced by a $k$-winners-take-all mechanism, in which exactly $k$ units out of $m$ are active for any given input, a construction that renders classical sparsity measures uninformative. Kurtosis and the Rolls \& Tov\'{e}e activity ratio \citep{rolls1995sparseness} were designed to quantify how closely the response distribution of a continuous-valued population approximates a sparse, near-binary pattern, and both degenerate when applied to a binary code with a fixed number of active bits, since two codes at the same sparsity level receive identical scores regardless of whether one places its active bits uniformly at random and the other reuses the same handful of bits for every input. What remains relevant for a binary code is therefore not the per-sample activation count, which is guaranteed by construction, but how evenly firing is distributed across the population over the full dataset.

We quantify this using the normalised Shannon entropy \citep{shannon1948mathematical} of the distribution formed by each bit's share of total firing activity across the dataset. Writing $o_j$ for the number of samples in which bit $j$ is active, each bit's share of total activity is $q_j = o_j / \sum_i o_i$, and the metric is

\begin{equation}
  \label{eq:entropy}
  H = \frac{-\sum_j q_j \log q_j}{\log m}
\end{equation}

dividing by $\log m$ to normalise to the $[0,1]$ interval. A value close to 1 indicates that all bits are used with comparable frequency across the dataset, while a value close to 0 indicates that firing is concentrated on a small subset of bits. This metric is equivalent to the codebook perplexity $m^H$ used in the product quantization \citep{jegou2011product} and VQ-VAE \citep{vanderoord2017neural} literatures.

\begin{figure}[htbp]
  \centering
  \includegraphics[width=\linewidth]{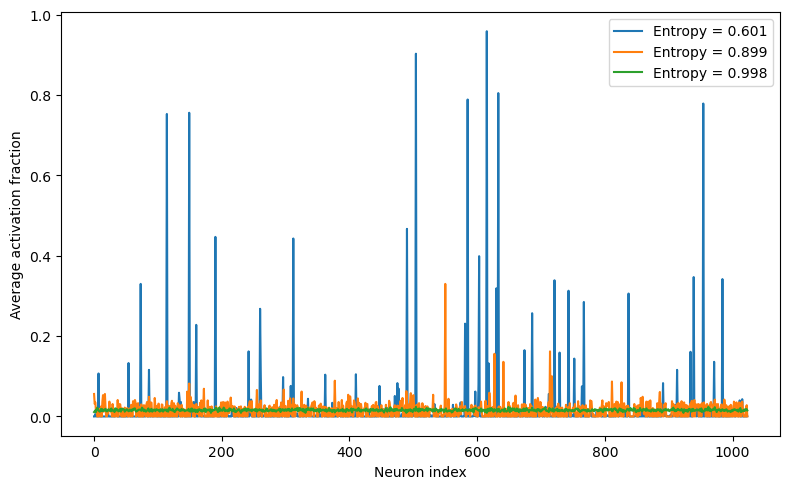}
  \caption{Per-neuron average activation fraction across 1000 samples with 1024-bit code and 16 active bits, for three example codes at increasing entropy. At low entropy (0.601), a small number of neurons fire far more often than the rest, producing sharp spikes against an otherwise near-silent population. At high entropy (0.899), this unevenness is much reduced. At entropy close to the ceiling (0.998), activation is almost perfectly flat across the entire population, closely matching what uniform usage of the code space looks like in practice.}
  \label{fig:entropy-activation-profile}
\end{figure}

\subsection{Information Preservation}
\label{sec:information-preservation}

This paper proposes to formalise information preservation, the second factor introduced in the Introduction, as similarity preservation, a specific form of information preservation concerned with relational information, namely which inputs belong together, rather than with the content of any single input in isolation. Similarity preservation is also the notion of information preservation most commonly sought in representation learning generally, where embeddings are explicitly designed so that proximity reflects similarity \citep{hadsell2006dimensionality,weinberger2009distance}, and it differs fundamentally from reconstruction fidelity, the classical target for autoencoders, which does not guarantee similarity preservation because a code can retain enough information to reconstruct an input precisely while still failing to place similar inputs near each other. Human perception illustrates this distinction, since people cannot recall the precise pixel values of a previously seen image yet reliably recognise that two images of the same handwritten digit under different rotations, translations, or stroke thicknesses belong to the same category, suggesting that similarity preservation is better captured by invariance to such perturbations than by fidelity to the original input \citep{dicarlo2007untangling}.

Similarity preservation admits no single canonical definition, in part because defining similarity even in the input space is a non-trivial problem. Euclidean distance is the most commonly used form of distance, but it is not invariant to translation, rotation, scale, or brightness, and distance measures engineered to be invariant to specific transformations only relocate the problem, since choosing which transformations to ignore is itself a decision about what similarity means. Increasing invariance to one kind of variation can moreover destroy distinctions the task genuinely depends on, since a distance measure made fully invariant to rotation would be unable to distinguish a six from a nine, one being the other rotated by 180 degrees. No single distance measure therefore resolves the definition of similarity in general. We use the MNIST dataset \citep{lecun1998gradient} for the comparisons that follow, a choice motivated in part by its standard preprocessing, which rescales each digit to a fixed bounding box and centres it by its centre of mass, thereby mitigating the most problematic sources of variation, specifically translation and scale, and making standard non-invariant distances such as Euclidean and Cosine a reasonable choice.

\paragraph{Class Overlap Separation.}
A direct way to measure similarity preservation using class labels is to compute the overlap between codes of the same class and subtract the overlap between codes of different classes, on the intuition that codes belonging to the same class should activate similar features and therefore share more active bits, while codes belonging to different classes should share fewer, and subtracting different-class overlap rather than reporting same-class overlap alone corrects for the baseline level of overlap that the code's overall sparsity would produce between any two codes regardless of class. The reported value is normalised by dividing by the number of active bits $k$. Formally, writing $S_c = \{i : y_i = c\}$ for the index set of class $c$ and $S_{\bar{c}} = \{i : y_i \neq c\}$ for all samples outside that class, and noting that for binary codes $\mathbf{z}_i^\top \mathbf{z}_j$ equals the number of bits simultaneously active in both codes,

\begin{equation}
  \label{eq:cos}
  \mathrm{COS} = \frac{1}{k} \left(
    \underset{\substack{i,j \in S_c \\ i \neq j}}{\mathbb{E}}\!\left[\mathbf{z}_i^\top \mathbf{z}_j\right]
    -
    \underset{\substack{i \in S_c \\ j \in S_{\bar{c}}}}{\mathbb{E}}\!\left[\mathbf{z}_i^\top \mathbf{z}_j\right]
  \right)
\end{equation}

averaged over classes $c = 1, \ldots, C$, where self-comparisons are excluded ($i \neq j$) and both expectations are estimated from the same number $P$ of randomly sampled pairs, ensuring the subtraction is not biased by group size.

\paragraph{Nearest-Class Impostor Margin.}
Class Overlap Separation pools every other class into a single average, which allows one badly confused pair of classes to hide behind several well-separated ones, a failure mode that would be unacceptable in downstream applications such as associative memory or classification regardless of how well every other class is handled. Replacing this average with a minimum closes this gap, since for each class the overlap with every other class is computed individually and the worst case, the nearest impostor, is taken rather than the average, so that a code in which fours and nines produce nearly identical activations is exposed by this margin even if every other digit pair is well separated. The reported metric is the per-class margin, same-class overlap minus overlap with the nearest impostor class, averaged across all classes and normalised by $k$. Writing $\bar{o}_{c,c'} = \frac{1}{|S_c||S_{c'}|} \sum_{i \in S_c,\, j \in S_{c'}} \mathbf{z}_i^\top \mathbf{z}_j$ for the mean overlap between classes $c$ and $c'$,

\begin{equation}
  \label{eq:nim}
  \mathrm{NIM} = \frac{1}{kC} \sum_{c=1}^{C} \left( \bar{o}_{c,c} - \max_{c' \neq c}\, \bar{o}_{c,c'} \right)
\end{equation}

\paragraph{kNN Accuracy.}
This metric belongs to a family of measures from the dimensionality-reduction literature concerned with local neighbourhood preservation, asking whether an embedding keeps each point's nearby neighbours nearby rather than testing global structure preservation by correlating full pairwise distance matrices. Established neighbourhood-preservation measures such as trustworthiness and continuity \citep{venna2006local} check this by verifying whether the same individual points remain each other's nearest neighbours across both spaces, weighting errors by how badly ranked the intruding or lost neighbour is, but kNN Accuracy replaces this identity-based check with a label-based one, comparing the fraction of each point's $K$ nearest neighbours that share its class label in code space against the same fraction in pixel space. This substitution is deliberate rather than a simplification, since an identity-based check would require treating pixel-space neighbour relationships as ground truth, an assumption already undermined by the invariance failures of Euclidean distance established above. The reported metric is this class-agreement rate computed directly in code space, with the pixel-space rate shown alongside as a reference baseline rather than subtracted, so that a value above the baseline indicates the code improves on the neighbourhood structure already present in the raw input, and a value below indicates that the encoding has discarded useful structure that pixel space provided for free. Letting $\mathcal{N}_K(i)$ denote the $K$ nearest neighbours of sample $i$ in code space by descending bit overlap,

\begin{equation}
  \label{eq:knn-acc}
  \mathrm{kNN\text{-}Acc} = \frac{1}{N} \sum_{i=1}^{N} \frac{\left|\{j \in \mathcal{N}_K(i) : y_j = y_i\}\right|}{K}
\end{equation}

\paragraph{Structure Preservation.}
The fourth metric abandons class labels entirely and instead asks whether the code preserves the full relational geometry of the input space, following the representational similarity analysis framework of Kriegeskorte et al.\ \citep{kriegeskorte2008representational}, itself grounded in Shepard's earlier work \citep{shepard1980multidimensional} relating an original distance structure to a derived one. Representational similarity analysis compares two representations of the same set of items by building a dissimilarity matrix within each representation separately and then correlating the two matrices, allowing a direct comparison even when the two representations share no common coordinate system, which is precisely the situation here since pixel space and a binary sparse code have no natural correspondence between individual dimensions. One matrix is built from pairwise distances in pixel space and one from pairwise bit overlap in code space, and this comparison is carried out at two levels, a class level using the mean distance and mean overlap between every pair of classes, and an instance level using distances and overlaps between individually sampled pairs of examples. Because no single distance measure resolves the definition of similarity, pixel-space distance is computed two ways, Euclidean distance, which is sensitive to each sample's overall magnitude, and Cosine similarity, which ignores magnitude and compares only direction, a genuinely different notion of similarity rather than a rescaling of the same one. At each level a Spearman correlation is reported, a rank correlation robust to the coarse discrete resolution of binary code overlap values. Unlike the other three metrics, this one makes no use of class labels at all, so its validity rests entirely on how good a proxy pixel-space distance is for genuine similarity, which is precisely why both Euclidean and Cosine distance are tested rather than only one. Writing $\mathbf{s}^{\mathrm{px}}$ for the vector of pairwise pixel-space similarities (negative Euclidean distance $s_{ij} = -\|x_i - x_j\|_2$, or Cosine similarity $s_{ij} = x_i^\top x_j / (\|x_i\|\|x_j\|)$) and $\mathbf{o}^{\mathrm{code}}$ for pairwise bit overlaps in code space,

\begin{equation}
  \label{eq:sp}
  \mathrm{SP} = \rho_S\!\left(\mathbf{s}^{\mathrm{px}},\, \mathbf{o}^{\mathrm{code}}\right)
\end{equation}

where $\rho_S$ is the Spearman rank correlation, computed separately at the instance level over individual sample pairs and at the class level over class-mean similarities and class-mean overlaps.

% ----------------------------------------------------------------
\section{Results}
% ----------------------------------------------------------------

\subsection{Experimental Setup}

Each model was trained on a subset of 1000 samples drawn from the MNIST dataset, with pixel values normalised to the $[0, 1]$ range before training. All trained models produced sparse codes of a fixed size, 1024 dimensions with 16 active bits, and the codes produced for these same 1000 training samples were then used to compute the entropy and similarity preservation metrics reported throughout this section. The models were evaluated across multiple configurations and random seeds, and the distributions shown in the figures that follow reflect this variability rather than single-run estimates.

\subsection{Individual Metric Analysis}
\label{sec:individual-metrics}

\begin{figure}[htbp]
  \centering
  \includegraphics[width=\linewidth]{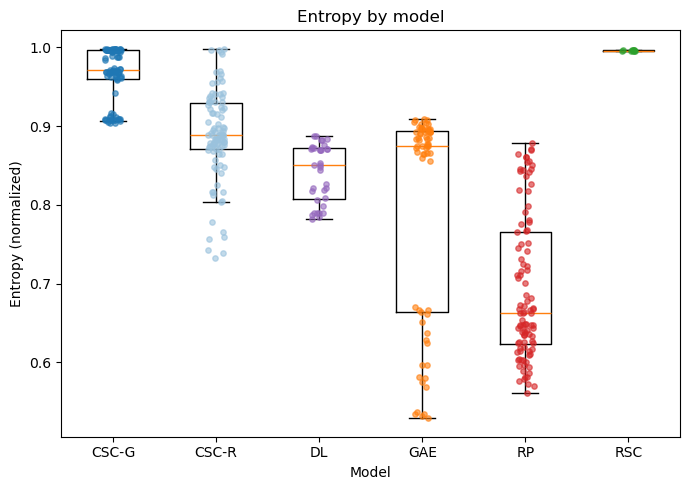}
  \caption{Entropy achieved by each method, against the random-sparse-code reference ceiling (RSC). CSC-G reaches entropy close to the RSC ceiling both consistently and tightly, while CSC-R reaches a similar ceiling only in its best configurations, with substantially more variation across the rest of its range. Dictionary Learning, GAE, and Random Projections all trail behind the ceiling, with different levels of deviation: Dictionary Learning is consistently high with little variation, while Random Projections shows much higher variation, and GAE displays a bimodal spread.}
  \label{fig:entropy_by_model}
\end{figure}

We begin by analysing sparsity in Fig.~\ref{fig:entropy_by_model}, comparing how closely each method's codes approach the uniform-usage ceiling represented by random sparse codes (RSC), the best-case scenario for entropy in which every bit is used with equal frequency. CSC is the only architecture that reaches entropy comparable to the RSC ceiling, and receptive field placement has a clear impact on how consistently it does so. CSC-G reaches the ceiling both tightly and reliably across configurations, while CSC-R reaches a comparable ceiling only in its best configurations, with substantially more variation across the rest of its range, plausibly because randomly placed receptive fields sometimes land in the periphery where a group's prototypes have less varied content to differentiate, leaving those groups with fewer distinct patterns across which to distribute winning bits. Although the distributions show real overlap between methods, particularly between CSC-R's lower range and those of DL and GAE, none of the other methods' maximum observed values approach the RSC ceiling that CSC's best runs reach under the studied parametrisations. Among the remaining methods, DL occupies a relatively narrow, consistently high band that never reaches the ceiling. RP spans a considerably wider range, suggesting that the particular random matrix drawn at initialisation has a substantial effect on how well the projection distributes activation across the code, but even its highest configurations fall short of the ceiling. The entropy distribution of GAE departs from all other methods in showing a bimodal structure, with configurations clustering into a lower band and a higher band rather than forming a continuous spread. We hypothesise that this separation is driven by the parameterisation of the KL weight $\lambda$, since configurations with lower $\lambda$ grant the model greater freedom to prioritise reconstruction fidelity over bit-usage uniformity, producing codes that fall into the lower entropy cluster, while configurations with higher $\lambda$ impose tighter uniformity constraints at the cost of reconstruction quality, pushing codes into the higher cluster.

\begin{figure}[htbp]
  \centering
  \includegraphics[width=\linewidth]{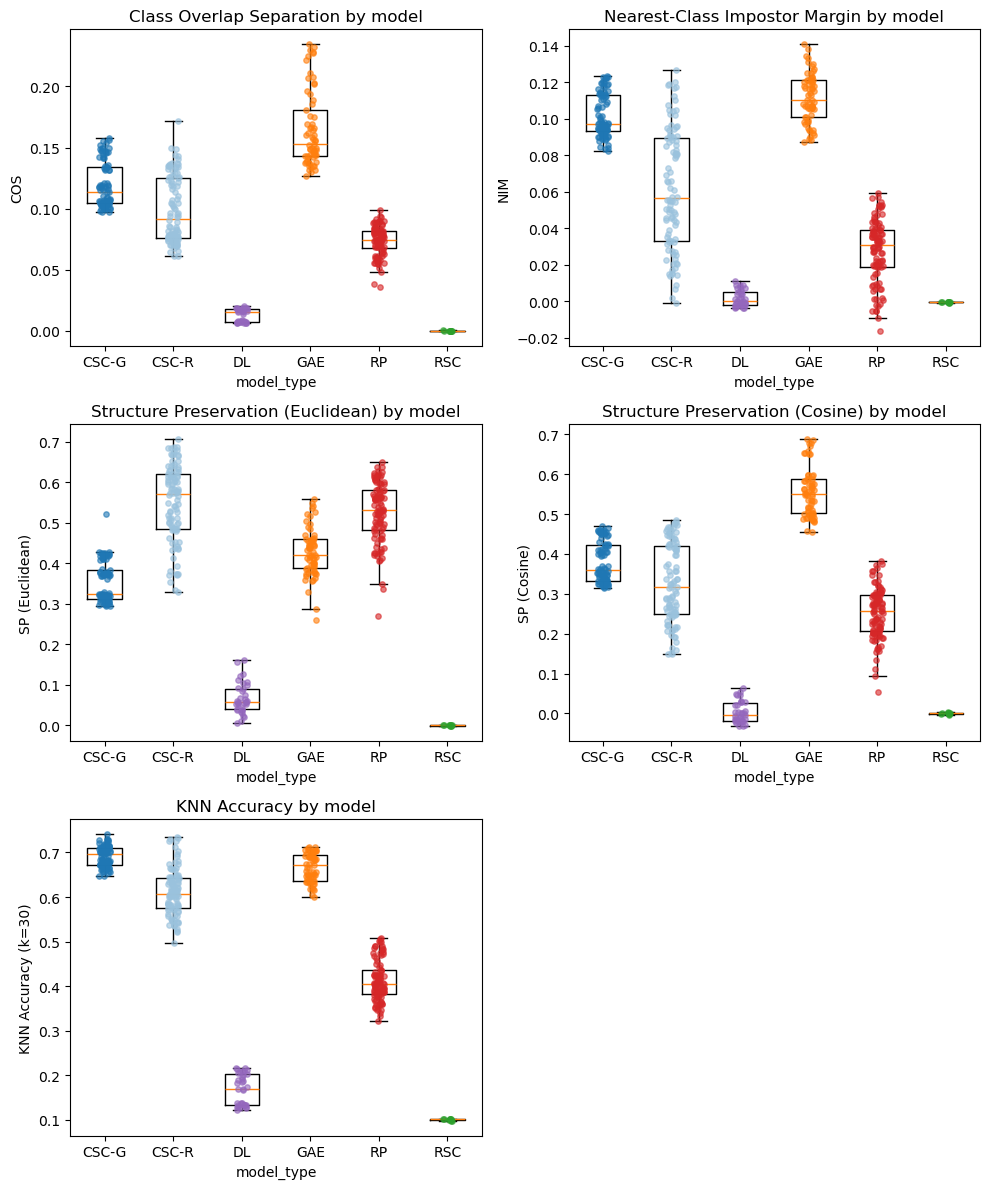}
  \caption{Per-model distribution of Class Overlap Separation, Nearest-Class Impostor Margin, Structure Preservation (Euclidean and Cosine), and kNN Accuracy, summarising the results above independently of entropy.}
  \label{fig:metrics_by_model}
\end{figure}

We now look at the results for similarity preservation considered independently of entropy in Fig.~\ref{fig:metrics_by_model}. RSC is retained in this comparison specifically to confirm that, as expected, codes generated with no reference to the input preserve no information on any metric, providing a lower bound against which all trained methods are compared. No single method outperforms every other on every metric when results are considered independently of entropy. DL performs worse than all other trained methods across all four metrics. This is likely a consequence of the $k$-WTA post-processing step applied to DL's continuous codes; unlike the other methods, which produce binary codes natively, DL optimises for reconstruction fidelity in continuous space and the subsequent hard thresholding discards much of the structure it learned, resulting in codes that are sparse by construction but retain little of the original information structure. GAE achieves the strongest typical performance on COS and NIM, with CSC-G competitive particularly on NIM. Both CSC and GAE perform well on kNN Accuracy relative to the other methods, though CSC-G stands out as the more consistent of the two, showing less variation across configurations and reaching slightly higher values than GAE. CSC-R and RP show a clear preference for preserving Euclidean structure over Cosine, while GAE shows the opposite preference, performing better on Cosine than Euclidean structure preservation, suggesting its learned representation is more sensitive to directional similarity than to differences in overall activation magnitude. CSC-G shows no strong preference in either direction, performing comparably on both Euclidean and Cosine structure preservation. These differing preferences make clear that the four metrics do not all capture the same dimension of similarity preservation, and that a method's relative standing depends meaningfully on which dimension is being measured.

\subsection{Relational Analysis}

\begin{figure}[htbp]
  \centering
  \includegraphics[width=0.75\linewidth]{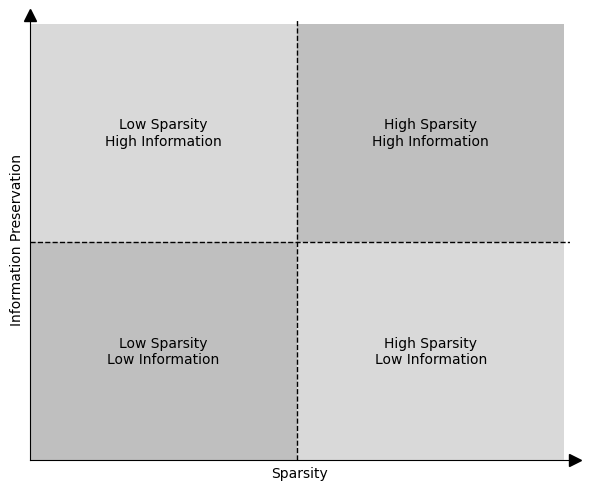}
  \caption{Conceptual framework for the results that follow, sparsity and information preservation define two axes along which each method's codes can be positioned. The high-sparsity, high-information quadrant (top right) is the desired outcome the central hypothesis holds is difficult to reach; the low-sparsity, low-information quadrant (bottom left) is trivial and uninteresting. The two off-diagonal quadrants represent the trade-off itself, sparsity gained at the cost of information, or information retained at the cost of sparsity.}
  \label{fig:quadrant-schematic}
\end{figure}

Examining sparsity and information preservation separately is not sufficient to understand how the two properties relate to each other, which is the central question motivating this paper's hypothesis, and the following analysis therefore considers both axes jointly, as illustrated in Fig.~\ref{fig:quadrant-schematic}.

Since sparsity and information preservation can each independently be high or low, jointly considering them naturally partitions the space of possible codes into four regions. A code that is both sparse and information-preserving occupies the most desirable region, which the paper's central hypothesis holds is difficult to reach, while a code that is neither is trivial and uninteresting, failing on both properties without offering any compensating advantage. The two remaining regions capture the trade-off directly, since a code that is sparse but fails to preserve information has paid the cost the hypothesis predicts, while a code that preserves information but is not sparse has only avoided that cost by giving up the property that motivated the comparison. The plots that follow adopt this layout, with sparsity measured by the entropy metric on the horizontal axis and information preservation measured by each of the metrics introduced in Section~\ref{sec:information-preservation} on the vertical axis.

\begin{figure}[htbp]
  \centering
  \includegraphics[width=\linewidth]{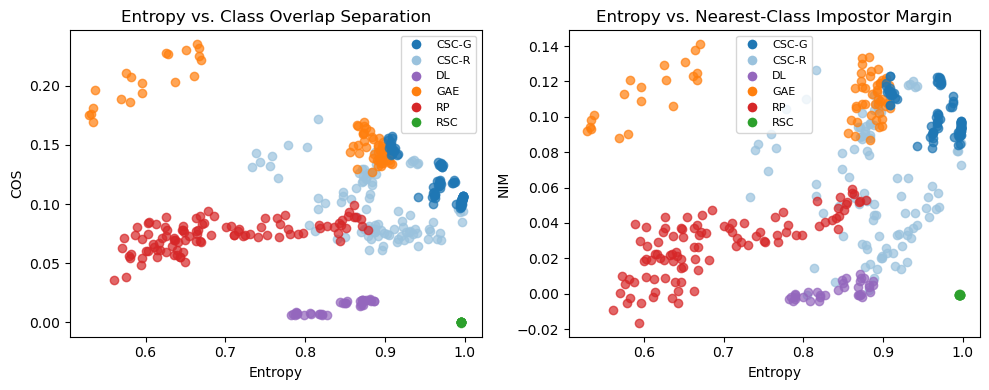}
  \caption{Entropy versus Class Overlap Separation and Nearest-Class Impostor Margin, across all evaluated configurations. GAE achieves the highest values on both metrics overall, though its configurations split into two clusters reflecting its bimodal entropy distribution, with the lower-entropy cluster reaching higher values on both similarity preservation metrics. DL and RP remain confined to comparatively low bands with little apparent dependence on entropy within their own ranges. Within the same entropy values, CSC-R and GAE perform relatively similarly on both COS and NIM, contrary to what the univariable analysis suggested, with CSC remaining able to push further beyond the 0.9 entropy threshold at a slight cost to similarity preservation.}
  \label{fig:entropy-class-metrics}
\end{figure}

In Fig.~\ref{fig:entropy-class-metrics} we examine both sparsity and our two initial supervised similarity preservation metrics jointly. The advantage GAE showed in the univariable analysis becomes less straightforward once entropy is taken into account. GAE's higher COS and NIM values are concentrated in its lower-entropy cluster, reflecting configurations where the KL weight $\lambda$ was set low enough to largely disregard entropy in favour of reconstruction fidelity. When comparing GAE and CSC-R within their shared entropy range, the performance gap narrows considerably, suggesting that GAE's apparent lead in Section~\ref{sec:individual-metrics} is driven primarily by those low-entropy configurations rather than by a uniform advantage across all conditions. DL and RP show little dependence on entropy within their own ranges, remaining confined to comparatively low bands throughout. CSC remains the only method capable of reaching entropy values above 0.9 while still maintaining meaningful COS and NIM values, though at a slight cost relative to its lower-entropy configurations.

\begin{figure}[htbp]
  \centering
  \includegraphics[width=\linewidth]{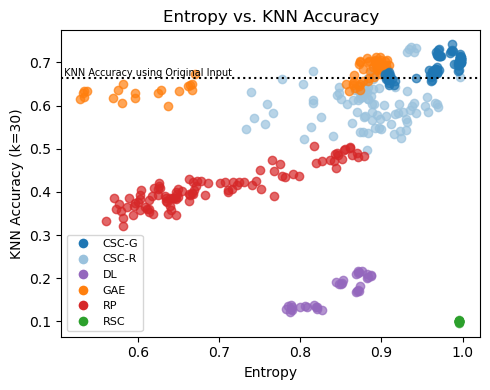}
  \caption{Entropy versus kNN Accuracy ($K=30$), across all evaluated configurations. CSC-G is the strongest performer overall, consistently at or above the pixel-space baseline. Contrary to expectation, kNN Accuracy appears to increase with entropy across most methods rather than decline.}
  \label{fig:entropy-knn}
\end{figure}

As seen in Fig.~\ref{fig:entropy-knn}, kNN Accuracy's positive relationship with entropy is consistent across most methods, suggesting that more uniform bit usage facilitates rather than hinders neighbourhood-based classification. CSC-G is the only method to sit consistently above the pixel-space baseline, indicating that its codes produce a neighbourhood structure genuinely better than that of the raw input. The remaining trained methods approach the baseline at their higher-entropy configurations, with Dictionary Learning sitting notably below it across its entire range.

\begin{figure}[htbp]
  \centering
  \includegraphics[width=\linewidth]{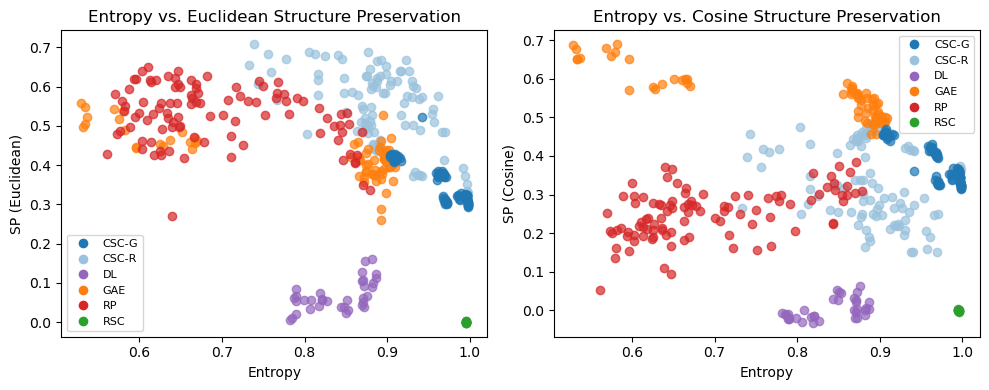}
  \caption{Structure Preservation against Entropy, computed separately using Euclidean (left) and Cosine (right) pixel-space distance.}
  \label{fig:entropy-structure}
\end{figure}

\begin{figure}[htbp]
  \centering
  \includegraphics[width=\linewidth]{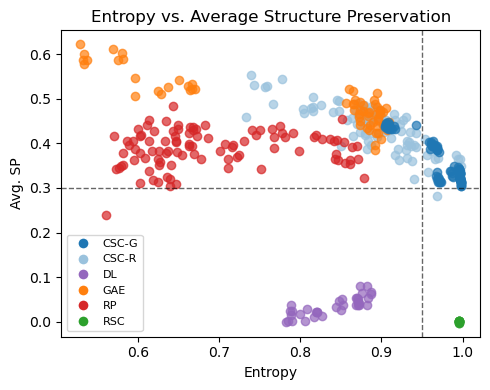}
  \caption{Entropy versus the average of the two Structure Preservation variants shown above. The dashed lines mark example thresholds for what counts as sufficiently high sparsity and information preservation, added to make the comparison with the quadrant framework introduced above easier to see. Under this choice of thresholds, only CSC reaches the desired quadrant, and only at the highest entropy values within its range.}
  \label{fig:entropy-avg-structure}
\end{figure}

As identified in Section~\ref{sec:individual-metrics}, some models show a clear preference for preserving one form of distance over the other, a pattern visible again here in Fig.~\ref{fig:entropy-structure}. To enable a fair comparison across models given this asymmetry, the Euclidean and Cosine Structure Preservation values are combined into a single averaged metric, arriving at Fig.~\ref{fig:entropy-avg-structure}.

Relating these results back to the quadrant framework, the dashed lines in the figure serve as worked examples of where such thresholds might be drawn, with the horizontal line set at a structure preservation of 0.3, the conventional threshold above which a correlation is no longer considered weak, and the vertical line set at entropy 0.95, a near-perfect level of uniformity close to the ideal represented by the RSC ceiling. Under these thresholds, most methods fall into the predicted categories, sacrificing sparsity in exchange for higher information preservation, with GAE as the clearest example. DL sits in the worst quadrant, weak on both axes, while RSC trivially falls into the expected quadrant of high sparsity paired with no information preservation. The most interesting result belongs to CSC, the only method able to reach a region where near-perfect sparsity coexists with a still-meaningful level of preserved information. The specific boundary of this quadrant is sensitive to where the thresholds are drawn, and the thresholds shown should be read as illustrative rather than principled, though the broader pattern they highlight, namely that CSC reaches higher entropy than any other method while retaining a non-trivial level of structure preservation, is visible directly in the scatter independently of where the lines are placed.

% ----------------------------------------------------------------
\section*{Conclusion}
% ----------------------------------------------------------------

Among the four families of sparse coding models compared in this work, Competitive Sparse Codes is the only one able to reach the near-maximal sparsity of an optimal random sparse code without collapsing information preservation to the level a random code exhibits, occupying a region of the sparsity--information-preservation space that no other method reaches, though its information preservation remains lower than that of models which sacrifice sparsity in exchange for higher similarity preservation. Within CSC, receptive field placement proves a consequential design choice: the Gaussian-centred variant (CSC-G) reaches near-maximal entropy reliably across configurations and leads all methods on kNN Accuracy, while the random placement variant (CSC-R) shows substantially more variation and weaker performance across the same metrics, suggesting that concentrating receptive fields on the informationally richer central region of the input has a meaningful effect on the quality of the resulting codes. The deep learning approach, by contrast, achieved the highest information preservation of all methods evaluated yet appears unable to exceed a certain level of sparsity regardless of how heavily the sparsity term is weighted in the loss function, confirming that the two properties resist simultaneous optimisation in different ways depending on the architecture.

The metrics introduced in this work help clarify the dynamics between sparsity and information preservation, and the results obtained through them confirm the open challenge framed at the beginning of this work, suggesting that approaches capable of satisfying both factors simultaneously and robustly likely remain to be found. This evaluation was carried out on a single, relatively simple dataset, and it remains an open question whether these findings generalise to more complex, higher-dimensional natural image datasets, though addressing this challenge will likely require further architectural innovation, with the biologically-inspired direction explored through Competitive Sparse Codes, local Hebbian learning combined with structured receptive fields, appearing the most promising avenue among the methods considered here. The evaluation framework introduced in this work is intended to serve as a reusable tool for measuring progress on this challenge as new architectures are proposed.

% ----------------------------------------------------------------
\subsection*{Acknowledgments}
% ----------------------------------------------------------------

This work was supported by national funds through Funda\c{c}\~{a}o para a Ci\^{e}ncia e Tecnologia , I.P. (FCT) under projects UID/50021/2025 and UID/PRR/50021/2025 and through a doctoral grant with reference 2023.01967.BD awarded to the first author. The funders had no role in study design, data collection and analysis, decision to publish, or preparation of the manuscript. The authors declare no conflicts of interest. Code and data for all the experiments can be obtained by email request to the first author.

\bibliographystyle{apalike}
\bibliography{references}

@article{lecun2015deep,
  author  = {LeCun, Yann and Bengio, Yoshua and Hinton, Geoffrey},
  title   = {Deep learning},
  journal = {Nature},
  year    = {2015},
  volume  = {521},
  number  = {7553},
  pages   = {436--444},
  doi     = {10.1038/nature14539}
}

@article{raichle2002appraising,
  author  = {Raichle, Marcus E. and Gusnard, Debra A.},
  title   = {Appraising the brain's energy budget},
  journal = {Proceedings of the National Academy of Sciences},
  year    = {2002},
  volume  = {99},
  number  = {16},
  pages   = {10237--10239},
  doi     = {10.1073/pnas.172399499}
}

@book{hebb1949organization,
  author    = {Hebb, Donald O.},
  title     = {The Organization of Behavior: A Neuropsychological Theory},
  year      = {1949},
  publisher = {Wiley},
  address   = {New York}
}

@article{rolls1995sparseness,
  author  = {Rolls, Edmund T. and Tovee, Martin J.},
  title   = {Sparseness of the neuronal representation of stimuli in the primate temporal visual cortex},
  journal = {Journal of Neurophysiology},
  year    = {1995},
  volume  = {73},
  number  = {2},
  pages   = {713--726},
  doi     = {10.1152/jn.1995.73.2.713}
}

@article{palm1980associative,
  author  = {Palm, G{\"u}nther},
  title   = {On associative memory},
  journal = {Biological Cybernetics},
  year    = {1980},
  volume  = {36},
  number  = {1},
  pages   = {19--31},
  doi     = {10.1007/BF00337019}
}

@article{willshaw1969nonholographic,
  author  = {Willshaw, D. J. and Buneman, O. P. and Longuet-Higgins, H. C.},
  title   = {Non-Holographic Associative Memory},
  journal = {Nature},
  year    = {1969},
  volume  = {222},
  pages   = {960--962},
  doi     = {10.1038/222960a0}
}

@article{olshausen1996emergence,
  author  = {Olshausen, Bruno A. and Field, David J.},
  title   = {Emergence of simple-cell receptive field properties by learning a sparse code for natural images},
  journal = {Nature},
  year    = {1996},
  volume  = {381},
  number  = {6583},
  pages   = {607--609},
  doi     = {10.1038/381607a0}
}

@inproceedings{jang2017categorical,
  author    = {Jang, Eric and Gu, Shixiang and Poole, Ben},
  title     = {Categorical Reparameterization with Gumbel-Softmax},
  booktitle = {International Conference on Learning Representations (ICLR)},
  year      = {2017}
}

@article{sacouto2023competitive,
  author  = {S{\'a}-Couto, Lu{\'i}s and Wichert, Andreas},
  title   = {Competitive learning to generate sparse representations for associative memory},
  journal = {Neural Networks},
  year    = {2023},
  volume  = {168},
  pages   = {32--43},
  doi     = {10.1016/j.neunet.2023.09.005}
}

@article{vinje2000sparse,
  author  = {Vinje, William E. and Gallant, Jack L.},
  title   = {Sparse Coding and Decorrelation in Primary Visual Cortex During Natural Vision},
  journal = {Science},
  year    = {2000},
  volume  = {287},
  number  = {5456},
  pages   = {1273--1276},
  doi     = {10.1126/science.287.5456.1273}
}

@incollection{barlow1959sensory,
  author    = {Barlow, Horace B.},
  title     = {Sensory mechanisms, the reduction of redundancy, and intelligence},
  booktitle = {The Mechanisation of Thought Processes},
  year      = {1959},
  publisher = {HMSO},
  address   = {London},
  note      = {National Physical Laboratory Symposium No. 10},
  pages     = {535--539}
}

@incollection{barlow1961possible,
  author    = {Barlow, Horace B.},
  title     = {Possible Principles Underlying the Transformation of Sensory Messages},
  booktitle = {Sensory Communication},
  editor    = {Rosenblith, Walter A.},
  year      = {1961},
  publisher = {MIT Press},
  address   = {Cambridge, MA},
  pages     = {217--234}
}

@article{atick1990towards,
  author  = {Atick, Joseph J. and Redlich, A. Norman},
  title   = {Towards a Theory of Early Visual Processing},
  journal = {Neural Computation},
  year    = {1990},
  volume  = {2},
  number  = {3},
  pages   = {308--320},
  doi     = {10.1162/neco.1990.2.3.308}
}

@article{atick1992retina,
  author  = {Atick, Joseph J. and Redlich, A. Norman},
  title   = {What Does the Retina Know about Natural Scenes?},
  journal = {Neural Computation},
  year    = {1992},
  volume  = {4},
  number  = {2},
  pages   = {196--210},
  doi     = {10.1162/neco.1992.4.2.196}
}

@article{field1994goal,
  author  = {Field, David J.},
  title   = {What Is the Goal of Sensory Coding?},
  journal = {Neural Computation},
  year    = {1994},
  volume  = {6},
  number  = {4},
  pages   = {559--601},
  doi     = {10.1162/neco.1994.6.4.559}
}

@article{laughlin1998metabolic,
  author  = {Laughlin, Simon B. and de Ruyter van Steveninck, Rob R. and Anderson, John C.},
  title   = {The metabolic cost of neural information},
  journal = {Nature Neuroscience},
  year    = {1998},
  volume  = {1},
  number  = {1},
  pages   = {36--41},
  doi     = {10.1038/236}
}

@article{attwell2001energy,
  author  = {Attwell, David and Laughlin, Simon B.},
  title   = {An Energy Budget for Signaling in the Grey Matter of the Brain},
  journal = {Journal of Cerebral Blood Flow \& Metabolism},
  year    = {2001},
  volume  = {21},
  number  = {10},
  pages   = {1133--1145},
  doi     = {10.1097/00004647-200110000-00001}
}

@article{lennie2003cost,
  author  = {Lennie, Peter},
  title   = {The Cost of Cortical Computation},
  journal = {Current Biology},
  year    = {2003},
  volume  = {13},
  number  = {6},
  pages   = {493--497},
  doi     = {10.1016/S0960-9822(03)00135-0}
}

@article{knoblauch2020iterative,
  author  = {Knoblauch, Andreas and Palm, G{\"u}nther},
  title   = {Iterative Retrieval and Block Coding in Autoassociative and Heteroassociative Memory},
  journal = {Neural Computation},
  year    = {2020},
  volume  = {32},
  number  = {1},
  pages   = {205--260},
  doi     = {10.1162/neco_a_01247}
}

@inproceedings{ravichandran2020learning,
  author    = {Ravichandran, Naresh Balaji and Lansner, Anders and Herman, Pawel},
  title     = {Learning Representations in {Bayesian} Confidence Propagating Neural Networks},
  booktitle = {2020 International Joint Conference on Neural Networks (IJCNN)},
  year      = {2020}
}

@inproceedings{hadsell2006dimensionality,
  author    = {Hadsell, Raia and Chopra, Sumit and LeCun, Yann},
  title     = {Dimensionality Reduction by Learning an Invariant Mapping},
  booktitle = {Proceedings of the IEEE Conference on Computer Vision and Pattern Recognition (CVPR)},
  year      = {2006},
  doi       = {10.1109/CVPR.2006.100}
}

@article{weinberger2009distance,
  author  = {Weinberger, Kilian Q. and Saul, Lawrence K.},
  title   = {Distance Metric Learning for Large Margin Nearest Neighbor Classification},
  journal = {Journal of Machine Learning Research},
  year    = {2009},
  volume  = {10},
  pages   = {207--244}
}

@article{lecun1998gradient,
  author  = {LeCun, Yann and Bottou, L{\'e}on and Bengio, Yoshua and Haffner, Patrick},
  title   = {Gradient-Based Learning Applied to Document Recognition},
  journal = {Proceedings of the IEEE},
  year    = {1998},
  volume  = {86},
  number  = {11},
  pages   = {2278--2324},
  doi     = {10.1109/5.726791}
}

@article{dicarlo2007untangling,
  author  = {DiCarlo, James J. and Cox, David D.},
  title   = {Untangling Invariant Object Recognition},
  journal = {Trends in Cognitive Sciences},
  year    = {2007},
  volume  = {11},
  number  = {8},
  pages   = {333--341},
  doi     = {10.1016/j.tics.2007.06.010}
}

@article{shannon1948mathematical,
  author  = {Shannon, Claude E.},
  title   = {A Mathematical Theory of Communication},
  journal = {Bell System Technical Journal},
  year    = {1948},
  volume  = {27},
  number  = {3},
  pages   = {379--423},
  doi     = {10.1002/j.1538-7305.1948.tb01338.x}
}

@article{venna2006local,
  author  = {Venna, Jarkko and Kaski, Samuel},
  title   = {Local multidimensional scaling},
  journal = {Neural Networks},
  year    = {2006},
  volume  = {19},
  number  = {6--7},
  pages   = {889--899},
  doi     = {10.1016/j.neunet.2006.05.014}
}

@article{kriegeskorte2008representational,
  author  = {Kriegeskorte, Nikolaus and Mur, Marieke and Bandettini, Peter A.},
  title   = {Representational Similarity Analysis -- Connecting the Branches of Systems Neuroscience},
  journal = {Frontiers in Systems Neuroscience},
  year    = {2008},
  volume  = {2},
  pages   = {4},
  doi     = {10.3389/neuro.06.004.2008}
}

@article{shepard1980multidimensional,
  author  = {Shepard, Roger N.},
  title   = {Multidimensional Scaling, Tree-Fitting, and Clustering},
  journal = {Science},
  year    = {1980},
  volume  = {210},
  number  = {4468},
  pages   = {390--398}
}

@inproceedings{gregor2010learning,
  author    = {Gregor, Karol and LeCun, Yann},
  title     = {Learning Fast Approximations of Sparse Coding},
  booktitle = {Proceedings of the 27th International Conference on Machine Learning (ICML)},
  year      = {2010},
  pages     = {399--406}
}

@incollection{johnson1984extensions,
  author    = {Johnson, William B. and Lindenstrauss, Joram},
  title     = {Extensions of {Lipschitz} Mappings into a {Hilbert} Space},
  booktitle = {Conference in Modern Analysis and Probability},
  series    = {Contemporary Mathematics},
  volume    = {26},
  year      = {1984},
  pages     = {189--206},
  publisher = {American Mathematical Society}
}

@inproceedings{makhzani2013ksparse,
  author    = {Makhzani, Alireza and Frey, Brendan},
  title     = {k-Sparse Autoencoders},
  booktitle = {International Conference on Learning Representations (ICLR)},
  year      = {2014}
}

@inproceedings{makhzani2015wta,
  author    = {Makhzani, Alireza and Frey, Brendan},
  title     = {Winner-Take-All Autoencoders},
  booktitle = {Advances in Neural Information Processing Systems (NeurIPS)},
  year      = {2015},
  pages     = {2791--2799}
}

@inproceedings{ravichandran2021brainlike,
  author    = {Ravichandran, Naresh Balaji and Lansner, Anders and Herman, Pawel},
  title     = {Brain-Like Approaches to Unsupervised Learning of Hidden Representations: {A} Comparative Study},
  booktitle = {Artificial Neural Networks and Machine Learning -- {ICANN} 2021},
  year      = {2021}
}

@article{dasgupta2017neural,
  author  = {Dasgupta, Sanjoy and Stevens, Charles F. and Navlakha, Saket},
  title   = {A neural algorithm for a fundamental computing problem},
  journal = {Science},
  year    = {2017},
  volume  = {358},
  number  = {6364},
  pages   = {793--796},
  doi     = {10.1126/science.aam9868}
}

@article{jegou2011product,
  author  = {J{\'e}gou, Herv{\'e} and Douze, Matthijs and Schmid, Cordelia},
  title   = {Product Quantization for Nearest Neighbor Search},
  journal = {IEEE Transactions on Pattern Analysis and Machine Intelligence},
  year    = {2011},
  volume  = {33},
  number  = {1},
  pages   = {117--128},
  doi     = {10.1109/TPAMI.2010.57}
}

@inproceedings{vanderoord2017neural,
  author    = {van den Oord, Aaron and Vinyals, Oriol and Kavukcuoglu, Koray},
  title     = {Neural Discrete Representation Learning},
  booktitle = {Advances in Neural Information Processing Systems (NeurIPS)},
  year      = {2017},
  pages     = {6306--6315}
}

\end{document}